\documentclass{INTERSPEECH2023}
\usepackage{multirow}
\usepackage{multicol}
\usepackage{arydshln}
\usepackage{algorithmic}
\usepackage{algorithm}
\usepackage{subfigure}
\usepackage{hyperref}
\interspeechcameraready

\title{Exploring Energy-based Language Models with Different Architectures and Training Methods for Speech Recognition}
\name{Hong Liu$^1$, Zhijian Ou$^2$, Third Author Name$^3$}
\name{Hong Liu$^{1, \dag}$, Zhaobiao Lv$^{2, \dag}$ \thanks{$^\dag$ Equal contribution. This work is supported by NSFC 61976122.}, Zhijian Ou$^{*,1}$\thanks{$^{*}$ Corresponding author: Zhijian Ou. The code is released at 
\url{https://github.com/thu-spmi/CAT/blob/master/docs/energy-based_LM_training.md}.
}, Wenbo Zhao$^{2}$, Qing Xiao$^{2}$
}
\address{
$^{1}$Speech Processing and Machine Intelligence (SPMI) Lab, Tsinghua University, Beijing, China \\
  $^{2}$China Unicom (Guangdong) Industrial Internet Co., Ltd.
}
\email{ liuhong21@mails.tsinghua.edu.cn, lvzb7@chinaunicom.cn, ozj@tsinghua.edu.cn, \\
\{zhaowb19, xiaoq17\}@chinaunicom.cn}

\begin{document}

\maketitle
 
\begin{abstract}
Energy-based language models (ELMs) parameterize an unnormalized distribution for natural sentences and are radically different from popular autoregressive language models (ALMs).
As an important application, ELMs have been successfully used as a means for calculating sentence scores in speech recognition, but they all use less-modern CNN or LSTM networks. 
The recent progress in Transformer networks and large pretrained models such as BERT and GPT2 opens new possibility to further advancing ELMs. 
In this paper, we explore different architectures of energy functions and different training methods to investigate the capabilities of ELMs in rescoring for speech recognition, all using large pretrained models as backbones.
% have a promising application prospect. However, compared with auto-regressive language models, the training and generation of ELMs is much more challenging, which leads to less extensive use of ELMs. 
% In this paper, we mainly explore the role of ELMs in the rescoring stage of speech recognition. We study ELMs from two aspects: the architecture and the training method. In terms of architecture, we summarize and improve a suite of past architectures for comparison. For training method, we try two mainstream methods, Maximum Likelihood Estimate (MLE) and Noise Contrastive Estimate (NCE) and some of their branches.
Extensive experiments are conducted on two datasets, AISHELL-1 and WenetSpeech. The results show that the best ELM achieves competitive results with the finetuned GPT2 and performs significantly better than the finetuned BERT. Further analysis show that the ELM obtains better confidence estimate performance than the finetuned GPT2.
\end{abstract}
\noindent\textbf{Index Terms}: energy-based language model, rescoring, speech recognition

\section{Introduction}
Energy-based language models (ELMs), as a class of energy-based models (EBMs) \cite{koller2009probabilistic,lecun2006tutorial}, parameterize an unnormalized distribution up to an unknown normalizing constant for natural sentences \cite{rosenfeld2001whole,trf,learning_trf}. ELMs are radically different from popular autoregressive language models (ALMs), which are locally normalized. 
Unfortunately, local normalization in ALMs brings some drawbacks, e.g., ALMs are prone to exposure bias \cite{wiseman2016sequence,ranzato2016sequence} and label bias \cite{lafferty2001conditional,andor2016globally}.
ELMs potentially address these issues, as they do not require any local normalization.
However, both exact computation of the normalizing constant and exact generation of samples from ELMs are generally intractable, which makes training especially difficult for ELMs.

% Different from the general self-normalized auto-regressive language models (ALMs), ELMs are theoretically globally normalized over the whole sentence space. Thus ELMs are naturally more suitable for scoring a sentence.

In recent years, there are encouraging progresses in both theories and applications of ELMs.
Applications of ELMs have covered computation of sentence likelihoods (up to a constant) \cite{trf,learning_trf,wang2017language,BinICASSP2018,DNCE,gao2020integrating}, text generation \cite{residual}, language model pretraining \cite{electric}, calibrated natural language understanding \cite{joint_ebm}, and so on.
% However, developing ELMs faces lots of challenges. On the one hand, the training and generation of ELMs are much more complicated than general language models due to the unknowability of the normalization constant. On the other hand, ELMs have few application scenarios at present and a thorough comparison among different ELMs is still lacking. 
% The focus of previous work on ELMs is different. 
As an important application, ELMs have been successfully used as a means for calculating sentence scores in automatic speech recognition (ASR).
\cite{trf,learning_trf} proposes trans-dimensional random field language model (TRF-LM) and applies to rescoring (i.e., reranking) of n-best lists for speech recognition. TRF-LMs outperform modified Kneser-Ney smoothing n-gram models \cite{chen1999empirical} when both using n-gram features.
Early ELMs are log-linear models \cite{rosenfeld2001whole,trf,learning_trf}. Later, ELMs using neural network based energy functions have been developed \cite{wang2017language,BinICASSP2018,DNCE}, outperforming ALMs with similar model sizes, but they all use old-fashioned CNN or LSTM networks. 
The recent progress in Transformer networks \cite{attention} and large pretrained models such as BERT \cite{bert} and GPT2 \cite{gpt2} opens new possibility to further advancing ELMs. 
% But they only use LSTM instead of the large pretrained language model as the backbone of the ELM, which leads to a mediocre performance. 
% \cite{electric} proposes an energy-based cloze model for pretraining and also applies it to the rescoring in ASR, but they did not finetuning the model to explore the ultimate performance of the model. 
% \cite{residual, joint_ebm} propose unique ELMs which however, are not used for rescoring in ASR.
In this paper, we explore different architectures of energy functions and different training methods to investigate the potential capabilities of ELMs in rescoring for speech recognition.

The architectures of energy functions in ELMs can be very flexibly defined. 
% We leave the detailed discussion to sections of Related Work and Method. 
In this work, we summarize and improve a suite of ELM architectures and name them SumTargetLogit, Hidden2Scalar, SumMaskedLogit and SumTokenLogit respectively.
% Systematic experiments are conducted to compare those architectures under different training methods. 
Model training of ELMs is challenging due to the intractable normalizing constant. 
% There are two mainstream training methods, maximum likelihood estimate (MLE) and noise contrastive estimate (NCE) \cite{nce}.
% In MLE, calculating gradients of the log likelihood usually resorts to Monte Carlo sampling methods. Two sampling methods are examined in this paper, importance sampling (IS) and Metropolis independent sampling (MIS).
% NCE, on the other hand, fitting unnormalized models by learning from distinguishing data samples and noise samples, where a noise distribution is required. Dynamic NCE (DNCE) \cite{DNCE} is an extension of NCE, which updates the noise distribution dynamically during training.
We leave detailed discussions to the sections of Related Work and Methods. 

Extensive experiments are conducted on two widely used Chinese speech recognition datasets, AISHELL-1 \cite{aishell} and WenetSpeech \cite{wenet}. We adopt large pretrained language models (PLMs) as the backbones of all energy models, noise models and proposal models in this work. We compare different combinations of architectures and training methods on these two datasets.
The results show that the best ELM achieves competitive results with the finetuned GPT2 and performs significantly better than the finetuned BERT. The advantage of ELM is more obvious on the large-scale WenetSpeech. Further analysis show that the ELM obtains better confidence estimate performance than the finetuned GPT2.
% By comparing different training methods, we find that DNCE outperforms NCE and MLE on different architectures.
% By comparing different architectures, we find that TRF-LM performs closely to the general Globally Normalized ELM (GN-ELM). Besides, the model architectures we summarized above have their own advantages and disadvantages under different training methods, but in general, the BERT-based architectures performs better.

\section{Related work}
% There has been quite a few works on ELM before. We will introduce them from two aspects: architectures and training methods. \\
\textbf{Architectures of ELMs}\\
One is generally free to choose the energy function, as long as it assigns a scalar energy to every sentence.
% and the distribution remains normalizable in principle.
TRF-LM \cite{DNCE} builds ELM in different dimensions according to sentence lengths, and directly sum the logit output from an ALM as energy, which corresponds to the SumTargetLogit architecture in this paper. Electric \cite{electric} is not strictly an ELM over sentences. It is in fact a cloze model, using  contextualized encoder outputs to define conditional energies. Electric leverages the pseudo-log-likelihood (PLL) \cite{pll} to score the sentence, which inspires the SumMaskedLogit and SumTokenLogit architectures. 
In \cite{joint_ebm}, three variants of energy functions are introduced. The first corresponds to the Hidden2Scalar architecture, and the latter two are based on a classification model, which is not relevant to the rescoring task in this paper.
\cite{residual} proposes a residual energy based model for conditional text generation, where a residual energy is defined on top of an ALM. 
% For instance, electric \cite{electric} focus on the improvement of pretraining cloze model, \cite{residual} devotes to text generation and \cite{joint_ebm} pays attention to better calibration of natural language understanding.
In addition to ELMs, there have also existed energy-based end-to-end speech recognition models for ASR, for which we refer readers to \cite{ctc-crf,variani2022global}.
\vspace{1em}\\
\textbf{Training methods for ELMs}\\
% MLE directly optimizes the expectation of log probability on data distribution, thus converts the processing of the normalization constant to sampling from the ELM. We can leverage Markov Chain Monte Carlo (MCMC) method to sample from the unnormalized ELM. MCMC includes many sub-methods, e.g., importance sampling \cite{IS}, where the proposed samples will be weighted and summed to calculate the expectation of the model distribution, metropolis hasting \cite{hastings1970monte} which uses accept/reject method to obtain samples from the model distribution, and metropolis independent sampling which is a special version of metropolis hasting.
% NCE \cite{nce} performs nonlinear logistic regression to discriminate between data samples and noise samples. The normalization constant of ELM can be treated as a normal parameter and updated together with the model parameters. The noise distribution is very important for the training stability and convergence of NCE. Dynamic NCE (DNCE) \cite{DNCE} is an extension of NCE, which uses an interpolation distribution between data and noise distributions to replace the original data distribution, and updates the noise distribution dynamically during training.
There are two mainstream training methods, maximum likelihood estimate (MLE) and noise contrastive estimate (NCE) \cite{nce}.
In MLE, calculating gradients of the log likelihood usually resorts to Monte Carlo sampling methods. 
Two widely-used classes of sampling methods are importance sampling (IS) and Markov Chain Monte Carlo (MCMC) \cite{liu2001monte}. MCMC covers a range of specific algorithms, e.g., Metropolis independent sampling (MIS) is a special instance, where the proposed move is generated independent of the previous state.
NCE, on the other hand, fitting unnormalized models by learning from distinguishing data samples and noise samples, where a noise distribution is required. Dynamic NCE (DNCE) \cite{DNCE} is an extension of NCE, which updates the noise distribution dynamically during training. 

\section{Methods}
\label{sec:method}
Let $x$ be a natural sentence (i.e., a token sequence). An energy-based language model (ELM) is defined as follows
\begin{equation}
\label{eq:elm}
    p_\theta(x)=\frac{\exp(-E_\theta(x))}{Z(\theta)}
\end{equation}
where $E_\theta(x)$ denotes an energy function with parameter $\theta$, $Z(\theta)=\sum_x \exp(-E_\theta(x))$ is the normalizing constant and $p_\theta(x)$ is the probability of sentence $x$. 
The design of $E_\theta(x)$ and the optimization of $\theta$ are the focus of this work.\\
\vspace{1em}\\
\textbf{TRF-LM}: \cite{trf} proposes trans-dimensional random field language model (TRF-LM), which builds energy models in different dimensions according to sentence lengths. Let $|x|$ be the length of sentence $x$, TRF-LM is defined as
\begin{equation}
\label{eq:trf}
    p_\theta(x)=\pi_{|x|} \frac{\exp(-E_\theta(x))}{Z_{|x|}(\theta)}
\end{equation}
where $Z_{|x|}(\theta)$ is the normalizing constant for length $|x|$. $\pi_{|x|}$ is the prior probability of length $|x|$, which is usually set as the empirical length probability, calculated from training data. 
The motivation of introducing length probabilities $\pi_{|x|}$ is that the empirical length probabilities can serve as a control device to improve sampling from multiple distributions over different lengths \cite{trf,learning_trf}.
To be differentiated from TRF-LM, the model in Eq.~\ref{eq:elm} is called globally-normalized ELM (\textbf{GN-ELM}).
% \textbf{Residual EBM}: \cite{residual} defines a residual energy on the basis of an ALM. The definition is
% \begin{equation}
%     p_\theta(x)=\frac{P_{LM}(x)\exp(-E_\theta(x))}{Z(\theta)}
% \end{equation}
% where $P_{LM}(x)$ is the probability of an ALM.
% Both variants above still belong to the scope of ELMs.
\subsection{Architectures of Energy Functions}
The architectures of energy functions in ELMs can be very flexibly defined. In the following, we summarize and introduce some architectures for ELMs.
Let $x=\{x_i\}_{i=1...|x|}$, where $x_i \in \{1,\cdots,V\}$ is the $i$-th token in $x$. 
$V$ denotes the size of token vocabulary. By abuse of notation, $x_i$  represents both the index of $x_i$ and the token itself.\\
\textbf{SumTargetLogit}: Similar to \cite{DNCE}, we borrow the architecture from ALMs.
Given history $x_{1:i-1}$, let the output logits to predict the next token be denoted by $f_{\theta}(x_{1:i-1})$, whose dimension is equal to $V$. The $k$-th logit is denoted by $f_{\theta}(x_{1:i-1})[k]$.
Then, the energy is defined as
\begin{equation}
\label{eq:sumtargetlogit}
    E_\theta(x)=-\sum_{i=1}^{|x|} f_{\theta}(x_{1:i-1})[x_i]
\end{equation}
% where $Id(x_{i+1})$ represents the index of $x_{i+1}$ in the vocabulary. 
This energy function sums the logits corresponding to the target token (next token) at each position, hence it is named by SumTargetLogit.
In contrast, the ALM applies local normalization (softmax) to the logits $f_{\theta}(x_{1:i-1})$ to obtain the conditional probability of $x_i$ given history $x_{1:i-1}$.
\\
\textbf{Hidden2Scalar}: The energy of SumTargetLogit is defined in uni-directional order like in ALMs.
More generally, like in \cite{wang2017language,residual,electric, joint_ebm}, we can use a bi-directional text encoder (e.g., BERT) to encode $x$ and we denote the encoder output (hidden vectors) by $\text{enc}_{\theta}(x)$. At position $i$, we have $\text{enc}_{\theta}(x)[i]$.
Then, the energy is defined as
\begin{equation}
\label{eq:hidden2scalar}
\vspace{-0.7em}
    E_\theta(x) = - \text{Linear} \left( \sum_{i=1}^{|x|} \text{enc}_\theta(x)[i] \right)
    % \vspace{-0.5em}
\end{equation}
% where $enc_\theta$ is an encoder which encodes the sentence into hidden vector sequence and 
where $\text{Linear}(\cdot)$ denotes a trainable linear layer whose output is a scalar.\\
\textbf{SumMaskedLogit}: For masked language model (MLM), e.g., BERT, pseudo-log-likelihood (PLL) is introduced for scroing sentences \cite{pll}. Inspired by this, we can define the energy function as follows:
\begin{equation}
\vspace{-0.5em}
    E_\theta(x)=-\sum_{i=1}^{|x|} g_\theta(\text{MASK}(x,i))[i][x_i]
    % \vspace{-0.5em}
\end{equation}
% Similar to SumTargetLogit, we take the logits of MLM as the energy. 
where $g_\theta$ denote the MLM, whose output, at each position, is the logits before softmax. 
$g_\theta(\text{MASK}(x,i))$ means masking the $i$-th token in $x$ and sending the masked sequence into the MLM for a forward pass. At position $i$, the logit corresponding to the masked token $x_i$ is denoted as $g_\theta(\text{MASK}(x,i))[i][x_i]$.
Notably, this architecture is much time-consuming than others since it requires $|x|$ forward passes to calculate the energy of one sentence, therefore we do not experiment with this architecture.\\
\textbf{SumTokenLogit}: To overcome the deficiency of SumMaskedLogit, we propose a simplication, i.e.,  omitting the masking step and feeding $x$ directly to the MLM, so that the logits at all positions can be calculated in parallel. The energy is defined as:
\begin{equation}
\label{eq:sumtokenlogit}
    E_\theta(x)=-\sum_{i=1}^{|x|}  g_\theta(x)[i][x_i]
\end{equation}

\subsection{Training Methods}
\subsubsection{Noise Contrastive Estimate}
\textbf{Noise Contrastive Estimate (NCE)} \cite{nce} optimizes the ELM by learning from discrimination between data samples and noise samples. Let $q_\phi$ be the noise distribution with parameter $\phi$, the NCE objective is formulated as:
\begin{displaymath}
\label{eq:NCE}
\resizebox{\linewidth}{!}{$\mathcal{J}_{\text{NCE}}(\theta)=\mathop{\mathbb{E}}\limits_{x \sim p_\text{data}} \log \frac{\hat{p_\theta}(x)}{\hat{p_\theta}(x)+\nu q_\phi(x)} + \nu \mathop{\mathbb{E}}\limits_{x \sim q_\phi} \log \frac{\nu q_\phi(x)}{\hat{p_\theta}(x)+\nu q_\phi(x)}$}
\end{displaymath}
where $\nu$ is the ratio between the noise prior and the data prior and $\hat{p_\theta}=\exp(-E_\theta(x))$ denotes the unnormalized probability. It is important that the noise distribution $q_\phi$ is close to the data distribution $p_\text{data}$ so that the binary classification task is sufficiently challenging for NCE to work \cite{nce}.\\
\textbf{Dynamic NCE (DNCE):} DNCE \cite{DNCE} was proposed with two motivations. One is to push the noise distribution to be close to the data distribution; the other is to prevent the model overfitting to the empirical distribution when the training data cannot represent the oracle data distribution. DNCE modifies NCE from the above two aspects, and we only adopt the first modification in this paper, that is performing maximum likelihood optimization of $q_\phi$'s parameter over the training data
\begin{equation}
\label{eq:dnce}
    \mathcal{J}_{\text{DNCE}}(\theta, \phi)=\mathcal{J}_{\text{NCE}}(\theta)+\mathbb{E}_{x\sim p_\text{data}} \log q_\phi(x)
\end{equation}
\subsubsection{Maximum Likelihood Estimate}
\label{sec:MLE}
% Maximum Likelihood Estimate (MLE) directly optimizes the expectation of the log probability over data distribution $p_d$, 
% The objective of maximum likelihood estimate (MLE) is:
% \begin{equation}
%     \mathcal{J}_{\text{MLE}}(\theta) =\mathbb{E}_{x \sim p_d} \log p_\theta(x)
% \end{equation}
The gradient of log likelihood in MLE learning of ELMs can be derived as follows:
\begin{equation}
\label{eq:MLE}
    \frac{\partial \mathcal{J}_{\text{MLE}}(\theta)} {\partial \theta}=-\mathbb{E}_{x \sim p_\text{data}} \left[\frac{\partial E_\theta(x)} {\partial \theta}\right] + \mathbb{E}_{x \sim p_\theta} \left[\frac{\partial E_\theta(x)} {\partial \theta}\right]
\end{equation}
The challenge is that the second expectation $ E_{x \sim p_\theta} [\frac{\partial E_\theta(x)} {\partial \theta}]$ requires sampling from the unnormalized ELM $p_\theta$. 
Similar to \cite{SNIS}, we compare two sampling approaches. Both methods need a proposal distribution $q_\phi$, which is implemented by an ALM in this paper. 
Note that the parameters of $q_\phi$ are also updated during training, similar to Eq.~\ref{eq:dnce}.\\
\textbf{Metropolis Independence Sampling (MIS)}: MIS is a special case of Metropolis-Hasting \cite{liu2001monte} and has been applied for ELM in \cite{wang2017language}. Algorithm~\ref{alg:MIS} shows the MIS details.
\begin{algorithm}[t]
	%\scriptsize
	\caption{Metropolis Independence Sampling in ELM.}
        \vspace{-0.5em}
 	\label{alg:MIS}
	\begin{algorithmic}
	    \REQUIRE A target distribution $p_\theta$, a proposal distribution $q_\phi$, iteration number $T$.
            \STATE Randomly initialize $x^{(0)}$;
            \FOR{$t$ =1 to $T$}
		\STATE Generate $x'$ from the proposal $q_\phi$;
            \STATE Accept $x^{(t)}=x'$ with probability \\ $\min\{1, \frac{p_\theta(x')q_\phi(x^{(t-1)})}{p_\theta(x^{(t-1)})q_\phi(x')}\}$, otherwise set $x^{(t)}=x^{(t-1)}$;
            \ENDFOR
        \STATE {\bf Return:} $\{x^{(1)},...,x^{(T)}\}$
	\end{algorithmic}
\end{algorithm}
In experiments, we run the Markov chain for $T$ steps, then use the samples $\{x^{(1)},...,x^{(T)}\}$ to approximate the expectation $ E_{x \sim p_\theta} [\frac{\partial E_\theta(x)} {\partial \theta}]$ in Eq.~\ref{eq:MLE} via Monte Carlo averaging.\\
% We'll obtain a realization of the Markov chain after the algorithm and $x^{(t)}$ can be regarded as a sample from $p_\theta$ if $t$ is big enough, which, can be used to calculate the expectation $ E_{x \sim p_\theta} [\frac{\partial E_\theta(x)} {\partial \theta}]$ in Eq.~\ref{eq:MLE}.\\
\textbf{Importance Sampling (IS)}: different from the accept/reject sampling mode in MIS, IS \cite{liu2001monte} computes an importance weight $w(x')=\frac{p_\theta(x')}{q_\phi(x')}$ for each proposed sample $x'$. For $N$ proposed samples $x'_1,...,x'_N$ from $q_\phi$, the second expectation in Eq.~\ref{eq:MLE} is estimated as
\begin{equation}
    \mathbb{E}_{x \sim p_\theta} \left[\frac{\partial E_\theta(x)} {\partial \theta} \right] \approx \frac{\sum_{i=1}^N w(x'_i)\frac{\partial E_\theta(x'_i)}{\partial \theta}}{\sum_{i=1}^N w(x'_i)}
\end{equation}
which theoretically is biased estimate.
% The estimate will converge to the true value as N approaches infinity.
Note that we restart the chain after each parameter update in applying MIS, hence its gradient estimate is also biased. One research question to be addressed in this work is to compare MLE based on MIS and IS with NCE and DNCE for learning ELMs.

\section{Experiments}
\subsection{Setup}
\textbf{Datasets.} We AISHELL-1 \cite{aishell} and WenetSpeech \cite{wenet} in experiments. AISHELL-1 is a 178-hour mandarin speech dataset and WenetSpeech is a 1000+ hours multi-domain transcribed Mandarin speech dataset.\\ %We first conduct a full experiment exploration on the small-scale AISHELL-1, and then run several representative experiments on WenetSpeech for validation. \\
\textbf{ASR Model.} The main task of ELM is to rescore the n-best list output from the first-pass ASR decoding. We interpolate the score of ASR model, the score of language model and the sentence length to get the final score. In this work, the ASR n-best lists are obtained from a RNN-T \cite{rnnt} model, where the encoder is a Conformer \cite{conformer} of 92M parameters. Note that the ASR model we use is rather competitive in terms of error rates of ``no LM'' on the two datasets in Table~\ref{tab:aishell} and \ref{tab:wenet}.\\
\textbf{Implementation Details.} We use Chinese GPT2 \cite{gpt2, chinesegpt2} as the ALM $f_\theta$ in Eq.~\ref{eq:sumtargetlogit} and Chinese BERT \cite{bert} as the encoder and MLM in Eq.~\ref{eq:hidden2scalar} and Eq.~\ref{eq:sumtokenlogit} respectively. Both pretrained models have 12 transformer layers with about 100M parameters. 
The noise distribution in NCE and the proposal distribution in MLE are both initialized from a finetuned GPT2, while the noise distribution in DNCE is initialized from a GPT2 without finetuning and is continuously optimized during training. We set the iteration number $T=256$ in MIS sampling in Algorithm~\ref{alg:MIS}.\\
\textbf{Test Details.} On AISHELL-1 test set, we use the ASR model trained on AISHELL-1 and that trained on WenetSpeech to generate n-best lists respectively, i.e., in-domain test and cross-domain test, the results of which are denoted as CER$_1$ and CER$_2$ respectively in Table~\ref{tab:aishell}. 
As for WenetSpeech, we only use the ASR model trained on itself to generate n-best lists on the two test sets of WenetSpeech, TEST-NET and TEST-MEETING. The results are also denoted as CER$_1$ and CER$_2$ respectively in Table~\ref{tab:wenet}.

\subsection{Results on AISHELL-1}
\label{sec:aishell}
Table \ref{tab:aishell} shows the Character Error Rate (CER) of different models on AISHELL-1 test set. 
The table is mainly divided into two parts. The first part is the rescoring results of non-energy models. Pretrained GPT2 and BERT represents that we rescore the n-best lists without any finetuning. Note that since BERT is not a traditional language model, we calculate the PLL mentioned above as the score of BERT.
The second part is the results of ELMs trained with different methods and architectures on transcripts of AISHELL-1, which are explained in Sec~\ref{sec:method}.

It can be seen that among all the ELMs, the GN-ELM with Hidden2Scalar architecture and trained with DNCE achieves the lowest CERs, which are on par with the best results achieved by the finetuned GPT2, showing the competitiveness of ELM for scoring sentence.
%We conduct matched pair test \cite{significance} between this ELM and the finetuned GPT2 and the $p$-value is 0.537 which is large enough to indicate that the difference between them is not significant. 
Besides, by observing and analyzing all the ELM experiments, we have the following conclusions:\\
\textbf{DNCE outperforms NCE.} We can see that the GN-ELMs trained with DNCE perform better than those trained with NCE under different architectures. This is not surprising since in DNCE, the binary classification challenge gradually increases with the optimization of noise model, which is more beneficial to optimizing ELM than a fixed noise model.\\
\textbf{GN-ELM and TRF-LM perform closely to each other.} When trained with DNCE, TRF-LM performs better than GN-ELM with SumTargetLogit, while it slightly underperforms GN-ELM with Hidden2Scalar and SumTokenLogit. \\
%We speculate that this is because the GN-ELM in Eq.~\ref{eq:elm} can automatically capture the dimension information in Eq~\ref{eq:trf}. 
\textbf{MLE underperforms NCE/DNCE.} Most of the results of GN-ELMs trained with MLE-IS and MLE-MIS are worse than those trained with NCE/DNCE. Moreover, we find that the training process of MLE is quite unstable and not easy to converge when the hyper-parameters are not appropriately tuned. We attribute this to the difficulty of sampling in the high dimensional space of ELM (or say in other words, obtaining unbiased gradient estimates) within acceptable number of IS/MIS steps.\\
\textbf{Bi-directional architectures perform better.} 
The bi-directional architectures (Hidden2Scalar and SumTokenLogit) based on BERT are generally better than the unidirectional architecture (SumTargetLogit) except DNCE (TRF-LM).
%None of those architectures is significantly better than the other two. However, SumTargetLogit performs worst in most cases except DNCE (TRF-LM).
% We think this is reasonable because Hidden2Scalar and SumTokenLogit are based on the bidirectional BERT, which can better capture the semantic information than the GPT-2 based SumTargetLogit. 
\begin{table}[t!]
  \caption{Rescoring results on AISHELL-1. CER$_1$ and CER$_2$ denote the Character Error Rate (CER) in in-domain test and cross-domain test respectively.}
  \label{tab:aishell}
  \centering
  \begin{tabular}{ll|cc}
    \hline
    Method & Architecture  & CER$_1$ &CER$_2$ \\
    \hline
    \multicolumn{2}{l|}{No LM} &4.76  & 5.14\\
    \multicolumn{2}{l|}{5-gram LM} &4.67 & 4.40 \\
     \multicolumn{2}{l|}{Pretrained GPT2} &3.22 &3.66 \\
    \multicolumn{2}{l|}{Pretrained BERT (PLL)} &3.29 &3.66 \\
    \multicolumn{2}{l|}{Finetuned GPT2} &3.11 &3.33\\
    \multicolumn{2}{l|}{Finetuned BERT (PLL)} &3.12 &3.47 \\
    \hline
    \multirow{4}{*}{NCE (GN-ELM)} &SumTargetLogit &3.32 &3.39\\
    &Hidden2Scalar &3.20 &3.36\\
    &SumTokenLogit &3.27 &3.43\\

    \hdashline
    \multirow{4}{*}{DNCE (GN-ELM)} &SumTargetLogit &3.25 &3.40\\
    &Hidden2Scalar &3.11 &3.34\\
    &SumTokenLogit &3.15 &3.43\\
    \hdashline
    \multirow{4}{*}{DNCE (TRF-LM)} &SumTargetLogit &3.11 &3.44\\
    &Hidden2Scalar &3.13 &3.39\\
    &SumTokenLogit &3.21 &3.47\\
    \hdashline
    \multirow{4}{*}{MLE-IS (GN-ELM)} &SumTargetLogit &3.42 &3.61\\
    &Hidden2Scalar &3.36 &3.48\\
    &SumTokenLogit &3.26 &3.41\\
    \hdashline
    \multirow{4}{*}{MLE-MIS (GN-ELM)} &SumTargetLogit &3.35 &3.59\\
    &Hidden2Scalar &3.26 &3.39\\
    &SumTokenLogit &3.25 &3.49\\
    \hline
  \end{tabular}
\end{table}

\subsection{Results on WenetSpeech}
According to the conclusions above, we only conduct experiments of GN-ELM and TRF-LM trained with DNCE with different architectures on transcripts of WenetSpeech. In fact, we also conducted MLE experiments on WenetSpeech but the training did not converge with the loss tending to be negative infinite.

Table~\ref{tab:wenet} shows the rescoring results on WenetSpeech. Different from the results on AISHELL-1, the GN-ELM with SumTokenLogit achieves the best results among all the ELMs and it even outperforms the finetuned GPT2. 
TRF-LMs do not achieve very good results on two test sets. Presumably, this is because the empirical length distribution obtained from the training data is not very applicable to the test set. 
Overall, the ELMs obtain the best performance upper on the large-scale WenetSpeech. This may reflect the benefit of ELMs in using bi-directional architecture and alleviating exposure bias and label bias.
\begin{table}[t]
  \caption{Rescoring results on WenetSpeech. CER$_1$ and CER$_2$ denote the CER in two test sets, TEST-NET and TEST-MEETING, respectively.}
  \vspace{-0.5em}
  \label{tab:wenet}
  \centering
  \resizebox{0.95\linewidth}{!}{
  \begin{tabular}{ll|cc}
    \hline
    Method & Architecture  & CER$_1$ &CER$_2$ \\
    \hline
    \multicolumn{2}{l|}{No LM} &9.69  & 17.91\\
     \multicolumn{2}{l|}{Pretrained GPT2} &9.10 &15.75 \\
    \multicolumn{2}{l|}{Pretrained BERT (PLL)} &9.07 &15.69 \\
    \multicolumn{2}{l|}{Finetuned GPT2} &8.82 &15.52\\
    \multicolumn{2}{l|}{Finetuned BERT (PLL)} &8.96 &15.55 \\
    \hline
    \multirow{4}{*}{DNCE (GN-ELM)} &SumTargetLogit &9.03 &16.02\\
    &Hidden2Scalar &8.98 &15.69\\
    &SumTokenLogit &8.81 &15.47\\
    \hdashline
    \multirow{4}{*}{DNCE (TRF-LM)} &SumTargetLogit &8.97 &15.77\\
    &Hidden2Scalar &8.95 &15.67\\
    &SumTokenLogit &9.00 &15.65\\
    \hline
  \end{tabular}}
\end{table}
\subsection{Analysis and Discussion}
\textbf{Significance Test.} To see whether the differences in Table~\ref{tab:aishell} and Table~\ref{tab:wenet} are significant, we conduct matched-pair significance test \cite{significance} for several pairs of experiments, whose $p$-values are shown in Table~\ref{tab:significance}. If we set the significance level $\alpha=0.05$, then all the experiment pairs with $p$-value less than 0.05 are considered be significantly different.
Main observations are as follows: a) The best GN-ELM (both in Table~\ref{tab:aishell} and Table~\ref{tab:wenet}) is significantly superior to the finetuned BERT in AISHELL-1 cross domain test and on the TEST-NET and TEST-MEETING of WenetSpeech, and it performs significantly better than the finetuned GPT2 on TEST-MEETING of WenetSpeech. 
Under other test situations, it achieves equally strong results as the finetuned GPT2 and BERT.
b) The best TRF-LM is on par with the best GN-ELM in AISHELL-1 in-domain test but it underperforms the GN-ELM under other test situations significantly.

\begin{figure}
	\centering

  \includegraphics[width=0.9\linewidth]{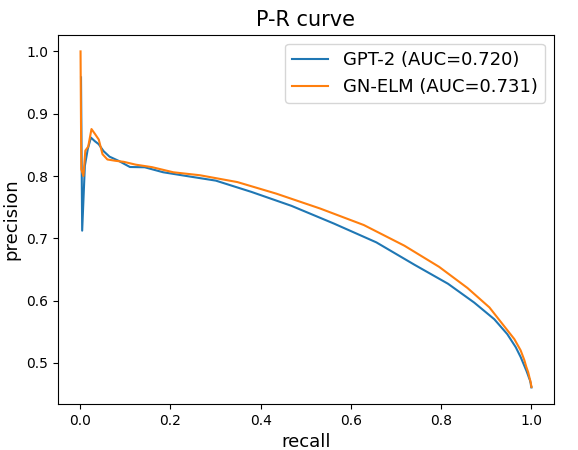}

		\caption{The confidence estimate performace of the finetuned GPT2 and the best ELM on the TEST-NET of WenetSpeech.}
		\label{fig:confidence}
  \end{figure}

\textbf{Confidence Estimate Performance.} 
The advantages of ELM also include confidence calibration \cite{joint_ebm, your}. Following \cite{confidence}, we plot the precision-recall curve by changing the confidence threshold and calculate the AUC. The results are shown in Figure~\ref{fig:confidence}. The best GN-ELM on WenetSpeech achieves a higher AUC, which illustrates that the GN-ELM has a better confidence estimate performance than the finetuned GPT2.
\vspace{-0.3em}

\begin{table}[t]
\caption{Matched pair test. The two $p$-values of each pair correspond to CER$_1$ and CER$_2$ in Table~\ref{tab:aishell} and Table~\ref{tab:wenet} respectively. A small $p$-value represents a more significant difference.}
    \centering
    \resizebox{\linewidth}{!}{
    \begin{tabular}{c|c|cc}
    \hline
      Dataset  & Model Pairs & \multicolumn{2}{c}{$p$-value} \\
      \hline
      \multirow{2}{*}{AISHELL-1} & Finetuned GPT2 & \multirow{2}{*}{0.979} &  \multirow{2}{*}{0.828} \\
      & DNCE (GN-ELM) + Hidden2Scalar\\
      \hline
      \multirow{2}{*}{AISHELL-1} & Finetuned BERT & \multirow{2}{*}{0.821} & \multirow{2}{*}{1e-5} \\
      & DNCE (GN-ELM) + Hidden2Scalar\\
      \hline
      \multirow{2}{*}{AISHELL-1} & DNCE (TRF-LM) + SumTargetLogit & \multirow{2}{*}{0.939} & \multirow{2}{*}{0.002} \\
      & DNCE (GN-ELM) + Hidden2Scalar\\
      \hline
      \multirow{2}{*}{WenetSpeech} & Finetuned GPT2 & \multirow{2}{*}{0.577} & \multirow{2}{*}{0.015} \\
      & DNCE (GN-ELM) + SumTokenLogit\\
      \hline
      \multirow{2}{*}{WenetSpeech} & Finetuned BERT & \multirow{2}{*}{1e-7} & \multirow{2}{*}{0.008} \\
      & DNCE (GN-ELM) + SumTokenLogit\\
      \hline
      \multirow{2}{*}{WenetSpeech} & DNCE (TRF-LM) + Hidden2Scalar  & \multirow{2}{*}{1e-7} & \multirow{2}{*}{1e-7} \\
      & DNCE (GN-ELM) + SumTokenLogit\\
    \hline
    \end{tabular}}
    \label{tab:significance}
    \vspace{-1em}
\end{table}

\section{Conclusions}
In this paper, we explore energy-based language models with different architectures and training methods for rescoring in ASR. We summarize and improve several architectures and examine four different training methods, all using large pretrained models as backbones. Experiments are conducted on two widely-used datasets and the results show that the best ELM can achieve competitive results with the finetuned GPT2 and significantly better results than the finetuned BERT. 
We hope these new findings would be helpful for future work to further explore ELMs.
% We hope also observe many interesting phenomena and conduct in-depth analysis, hoping to provide reference for future work.
%\lipsum[66]

\bibliographystyle{IEEEtran}
\bibliography{mybib}

\begin{thebibliography}{10}
\providecommand{\url}[1]{#1}
\csname url@samestyle\endcsname
\providecommand{\newblock}{\relax}
\providecommand{\bibinfo}[2]{#2}
\providecommand{\BIBentrySTDinterwordspacing}{\spaceskip=0pt\relax}
\providecommand{\BIBentryALTinterwordstretchfactor}{4}
\providecommand{\BIBentryALTinterwordspacing}{\spaceskip=\fontdimen2\font plus
\BIBentryALTinterwordstretchfactor\fontdimen3\font minus
  \fontdimen4\font\relax}
\providecommand{\BIBforeignlanguage}[2]{{%
\expandafter\ifx\csname l@#1\endcsname\relax
\typeout{** WARNING: IEEEtran.bst: No hyphenation pattern has been}%
\typeout{** loaded for the language `#1'. Using the pattern for}%
\typeout{** the default language instead.}%
\else
\language=\csname l@#1\endcsname
\fi
#2}}
\providecommand{\BIBdecl}{\relax}
\BIBdecl

\bibitem{koller2009probabilistic}
D.~Koller and N.~Friedman, \emph{Probabilistic graphical models: principles and
  techniques}.\hskip 1em plus 0.5em minus 0.4em\relax MIT press, 2009.

\bibitem{lecun2006tutorial}
Y.~LeCun, S.~Chopra, R.~Hadsell, M.~Ranzato, and F.~Huang, ``A tutorial on
  energy-based learning,'' \emph{Predicting structured data}, vol.~1, no.~0,
  2006.

\bibitem{rosenfeld2001whole}
R.~Rosenfeld, S.~F. Chen, and X.~Zhu, ``Whole-sentence exponential language
  models: a vehicle for linguistic-statistical integration,'' \emph{Computer
  Speech \& Language}, vol.~15, pp. 55--73, 2001.

\bibitem{trf}
B.~Wang, Z.~Ou, and Z.~Tan, ``Trans-dimensional random fields for language
  modeling,'' in \emph{Proceedings of the 53rd Annual Meeting of the
  Association for Computational Linguistics and the 7th International Joint
  Conference on Natural Language Processing (Volume 1: Long Papers)}, 2015, pp.
  785--794.

\bibitem{learning_trf}
B.~Wang, Z.~Ou, and Z.~Tan, ``Learning trans-dimensional random fields with
  applications to language modeling,'' \emph{IEEE transactions on pattern
  analysis and machine intelligence}, vol.~40, no.~4, pp. 876--890, 2017.

\bibitem{wiseman2016sequence}
S.~Wiseman and A.~M. Rush, ``Sequence-to-sequence learning as beam-search
  optimization,'' in \emph{EMNLP}, 2016.

\bibitem{ranzato2016sequence}
M.~Ranzato, S.~Chopra, M.~Auli, and W.~Zaremba, ``Sequence level training with
  recurrent neural networks,'' in \emph{4th International Conference on
  Learning Representations (ICLR)}, 2016.

\bibitem{lafferty2001conditional}
J.~Lafferty, A.~McCallum, and F.~C. Pereira, ``Conditional random fields:
  Probabilistic models for segmenting and labeling sequence data,'' in
  \emph{International conference on Machine learning (ICML)}, 2001.

\bibitem{andor2016globally}
D.~Andor, C.~Alberti, D.~Weiss, A.~Severyn, A.~Presta, K.~Ganchev, S.~Petrov,
  and M.~Collins, ``Globally normalized transition-based neural networks,'' in
  \emph{Proceedings of the 54th Annual Meeting of the Association for
  Computational Linguistics (Volume 1: Long Papers)}, 2016, pp. 2442--2452.

\bibitem{wang2017language}
B.~Wang and Z.~Ou, ``Language modeling with neural trans-dimensional random
  fields,'' in \emph{IEEE Automatic Speech Recognition and Understanding
  Workshop (ASRU)}.\hskip 1em plus 0.5em minus 0.4em\relax IEEE, 2017, pp.
  294--300.

\bibitem{BinICASSP2018}
B.~Wang and Z.~Ou, ``Learning neural trans-dimensional random field language
  models with noise-contrastive estimation,'' in \emph{International Conference
  on Acoustics, Speech, and Signal Processing (ICASSP)}, 2018.

\bibitem{DNCE}
B.~Wang and Z.~Ou, ``Improved training of neural trans-dimensional random field
  language models with dynamic noise-contrastive estimation,'' in \emph{2018
  IEEE Spoken Language Technology Workshop (SLT)}, 2018, pp. 70--76.

\bibitem{gao2020integrating}
S.~Gao, Z.~Ou, W.~Yang, and H.~Xu, ``Integrating discrete and neural features
  via mixed-feature trans-dimensional random field language models,'' in
  \emph{International Conference on Acoustics, Speech, and Signal Processing
  (ICASSP)}, 2020.

\bibitem{residual}
Y.~Deng, A.~Bakhtin, M.~Ott, A.~Szlam, and M.~Ranzato, ``Residual energy-based
  models for text generation,'' \emph{arXiv preprint arXiv:2004.11714}, 2020.

\bibitem{electric}
K.~Clark, M.-T. Luong, Q.~Le, and C.~D. Manning, ``Pre-training transformers as
  energy-based cloze models,'' in \emph{Proceedings of the 2020 Conference on
  Empirical Methods in Natural Language Processing (EMNLP)}, 2020, pp.
  285--294.

\bibitem{joint_ebm}
T.~He, B.~McCann, C.~Xiong, and E.~Hosseini-Asl, ``Joint energy-based model
  training for better calibrated natural language understanding models,'' in
  \emph{Proceedings of the 16th Conference of the European Chapter of the
  Association for Computational Linguistics: Main Volume}, 2021, pp.
  1754--1761.

\bibitem{chen1999empirical}
S.~F. Chen and J.~Goodman, ``An empirical study of smoothing techniques for
  language modeling,'' \emph{Computer Speech \& Language}, vol.~13, no.~4, pp.
  359--394, 1999.

\bibitem{attention}
A.~Vaswani, N.~Shazeer, N.~Parmar, J.~Uszkoreit, L.~Jones, A.~N. Gomez,
  {\L}.~Kaiser, and I.~Polosukhin, ``Attention is all you need,''
  \emph{Advances in neural information processing systems}, vol.~30, 2017.

\bibitem{bert}
J.~Devlin, M.-W. Chang, K.~Lee, and K.~Toutanova, ``Bert: Pre-training of deep
  bidirectional transformers for language understanding,'' \emph{arXiv preprint
  arXiv:1810.04805}, 2018.

\bibitem{gpt2}
A.~Radford, J.~Wu, R.~Child, D.~Luan, D.~Amodei, I.~Sutskever \emph{et~al.},
  ``Language models are unsupervised multitask learners,'' \emph{OpenAI blog},
  vol.~1, no.~8, p.~9, 2019.

\bibitem{aishell}
H.~Bu, J.~Du, X.~Na, B.~Wu, and H.~Zheng, ``Aishell-1: An open-source mandarin
  speech corpus and a speech recognition baseline,'' in \emph{2017 20th
  Conference of the Oriental Chapter of the International Coordinating
  Committee on Speech Databases and Speech I/O Systems and Assessment
  (O-COCOSDA)}, 2017, pp. 1--5.

\bibitem{wenet}
B.~Zhang, H.~Lv, P.~Guo, Q.~Shao, C.~Yang, L.~Xie, X.~Xu, H.~Bu, X.~Chen,
  C.~Zeng, D.~Wu, and Z.~Peng, ``Wenetspeech: A 10000+ hours multi-domain
  mandarin corpus for speech recognition,'' in \emph{International Conference
  on Acoustics, Speech, and Signal Processing (ICASSP)}, 2022, pp. 6182--6186.

\bibitem{pll}
A.~Wang and K.~Cho, ``{BERT} has a mouth, and it must speak: {BERT} as a
  {M}arkov random field language model,'' in \emph{Proceedings of the Workshop
  on Methods for Optimizing and Evaluating Neural Language Generation}, 2019,
  pp. 30--36.

\bibitem{ctc-crf}
H.~Xiang and Z.~Ou, ``Crf-based single-stage acoustic modeling with ctc
  topology,'' in \emph{IEEE International Conference on Acoustics, Speech and
  Signal Processing (ICASSP)}, 2019.

\bibitem{variani2022global}
E.~Variani, K.~Wu, M.~D. Riley, D.~Rybach, M.~Shannon, and C.~Allauzen,
  ``Global normalization for streaming speech recognition in a modular
  framework,'' \emph{Advances in Neural Information Processing Systems},
  vol.~35, pp. 4257--4269, 2022.

\bibitem{nce}
M.~Gutmann and A.~Hyv{\"a}rinen, ``Noise-contrastive estimation: A new
  estimation principle for unnormalized statistical models,'' in
  \emph{Proceedings of the thirteenth international conference on artificial
  intelligence and statistics}, 2010, pp. 297--304.

\bibitem{liu2001monte}
J.~S. Liu, \emph{Monte Carlo strategies in scientific computing}.\hskip 1em
  plus 0.5em minus 0.4em\relax Springer, 2001.

\bibitem{SNIS}
T.~Parshakova, J.-M. Andreoli, and M.~Dymetman, ``Global autoregressive models
  for data-efficient sequence learning,'' in \emph{Proceedings of the 23rd
  Conference on Computational Natural Language Learning (CoNLL)}, 2019, pp.
  900--909.

\bibitem{rnnt}
A.~Graves, ``Sequence transduction with recurrent neural networks,''
  \emph{arXiv preprint arXiv:1211.3711}, 2012.

\bibitem{conformer}
A.~Gulati, J.~Qin, C.-C. Chiu, N.~Parmar, Y.~Zhang, J.~Yu, W.~Han, S.~Wang,
  Z.~Zhang, Y.~Wu \emph{et~al.}, ``Conformer: Convolution-augmented transformer
  for speech recognition,'' \emph{arXiv preprint arXiv:2005.08100}, 2020.

\bibitem{chinesegpt2}
Z.~Zhao, H.~Chen, J.~Zhang, X.~Zhao, T.~Liu, W.~Lu, X.~Chen, H.~Deng, Q.~Ju,
  and X.~Du, ``Uer: An open-source toolkit for pre-training models,''
  \emph{EMNLP-IJCNLP 2019}, p. 241, 2019.

\bibitem{significance}
L.~Gillick and S.~J. Cox, ``Some statistical issues in the comparison of speech
  recognition algorithms,'' in \emph{International Conference on Acoustics,
  Speech, and Signal Processing (ICASSP)}, 1989, pp. 532--535.

\bibitem{your}
W.~Grathwohl, K.-C. Wang, J.-H. Jacobsen, D.~Duvenaud, M.~Norouzi, and
  K.~Swersky, ``Your classifier is secretly an energy based model and you
  should treat it like one,'' \emph{arXiv preprint arXiv:1912.03263}, 2019.

\bibitem{confidence}
Q.~Li, Y.~Zhang, D.~Qiu, Y.~He, L.~Cao, and P.~C. Woodland, ``Improving
  confidence estimation on out-of-domain data for end-to-end speech
  recognition,'' in \emph{International Conference on Acoustics, Speech, and
  Signal Processing (ICASSP)}, 2022, pp. 6537--6541.

\end{thebibliography}

\end{document}